\documentclass[10pt,twocolumn,letterpaper]{article}

\usepackage{wacv}
\usepackage{times}
\usepackage{epsfig}
\usepackage{graphicx}
\usepackage{amsmath}
\usepackage{amssymb}
\usepackage{booktabs}
\usepackage{lineno}

\usepackage{caption}
\usepackage{subcaption}

\usepackage{color}
\usepackage{xcolor}
\usepackage{soul}
\usepackage[normalem]{ulem}

\def\mypar#1{\vspace{0.15cm}\noindent{\bf #1.}}

\IfFileExists{darkmode.tex}{
    \usepackage{pagecolor}
    \definecolor{myfg}{gray}{0.94} 
    \definecolor{mybg}{gray}{0}
    \input{darkmode.tex} 
    \pagecolor{mybg}
    \color{myfg}
}{} 

\usepackage{amsmath,amssymb,amsbsy,xspace}

\def\<{\langle}
\def\>{\rangle}

\makeatletter
\DeclareRobustCommand\onedot{\futurelet\@let@token\@onedot}
\def\@onedot{\ifx\@let@token.\else.\null\fi\xspace}

\def\eg{\emph{e.g}\onedot} 
\def\ie{\emph{i.e}\onedot}

\makeatother

\def\wacvPaperID{421} 

\wacvfinalcopy 

\ifwacvfinal
\fi

\ifwacvfinal
\usepackage[breaklinks=true,bookmarks=false]{hyperref}
\else
\usepackage[pagebackref=true,breaklinks=true,colorlinks,bookmarks=false]{hyperref}
\fi

\usepackage{authblk}
\makeatletter
\renewcommand\AB@affilsepx{, \protect\Affilfont}
\makeatother

\begin{document}

\title{MaskSplit: Self-supervised Meta-learning for Few-shot Semantic Segmentation}

\author[1]{Mustafa\,Sercan\,Amac\thanks{Equal contribution. Listing order is random.}}
\newcommand\CoAuthorMark{\footnotemark[\arabic{footnote}]}
\author[2]{Ahmet\,Sencan\protect\CoAuthorMark}
\author[2]{Orhun Bugra\,Baran}
\author[3]{Nazli\,Ikizler-Cinbis}
\author[2]{Ramazan\,Gokberk\,Cinbis}
\affil[1]{\small MonolithAI} 
\affil[2]{\small Department of Computer Engineering, Middle East Technical University, Ankara, Turkey}
\affil[3]{\small Department of Computer Engineering, Hacettepe University, Ankara, Turkey}
\affil[ ]{\textit\small {sercanamac@gmail.com, \{sencan.ahmet,bbaran,gcinbis\}@metu.edu.tr, nazli@cs.hacettepe.edu.tr}}

\maketitle

\begin{abstract}
Just like other few-shot learning problems, few-shot segmentation aims to minimize the need for manual annotation, which is particularly costly in segmentation tasks. Even though the few-shot setting reduces this cost for novel test classes, there is still a need to annotate the training data. 
To alleviate this need, we propose a self-supervised training approach for learning few-shot segmentation models. We first use unsupervised saliency estimation to obtain pseudo-masks on images. We then train a simple prototype based model over different splits of pseudo masks and augmentations of images. Our extensive experiments show that the proposed approach achieves promising results, highlighting the potential of self-supervised training. To the best of our knowledge this is the first work that addresses unsupervised few-shot segmentation problem on natural images.

\end{abstract}

\section{Introduction}

Semantic segmentation is the task of assigning labels to pixels of a given image. There has
been tremendous progress in semantic segmentation with the developments in architectures
\cite{fcn,parsenet,unet,refinenet,deeplabv3,contextencoding}.  However, these approaches typically
require large amounts of training data for each class of interest to achieve accurate results.
The manual effort needed for collecting segmentation annotations greatly limits the scalability of such 
supervised approaches.

Aiming to mimic the human ability to recognize and segment novel object classes with just a few
examples, few-shot semantic segmentation has gained popularity over the past few years. In contrast
to supervised approaches, {\em few-shot semantic segmentation} (FSS) aims to estimate the mask for
an input image, \ie the {\em query} image, with the help of just a few training images and their
groundtruth masks, \ie the {\em support} samples.

\begin{figure}[t]
  \includegraphics[width=\linewidth]{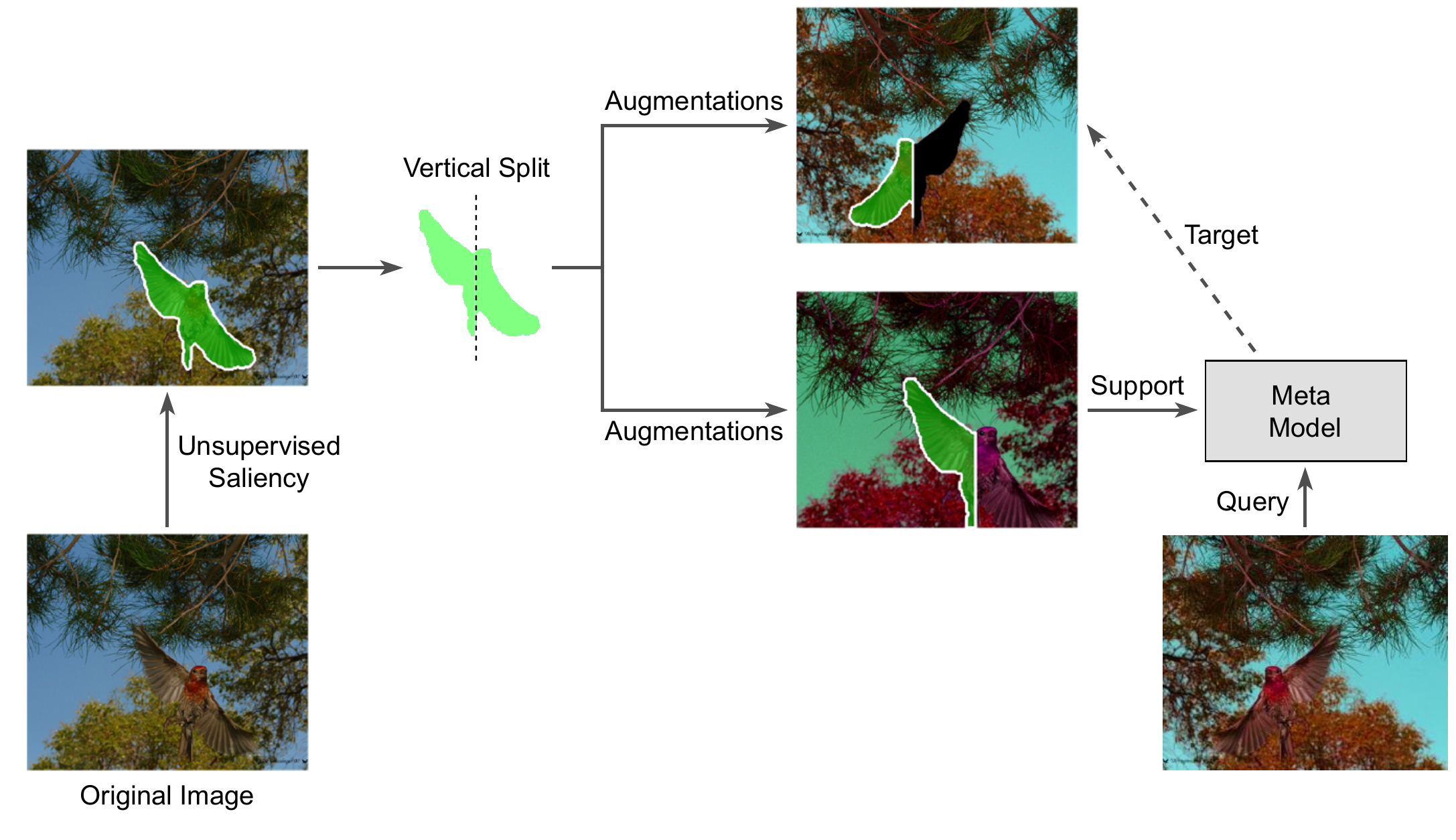}
    \caption{Illustration of the proposed self-supervised meta-learning framework. We split saliency masks to create artificial support and query training pairs, using which we train the few-shot segmentation model. Thanks to this self-supervised learning setup, no manual segmentation annotations are needed during training. }
  \label{fig:intro}
  \vspace{-2mm}
\end{figure}

The recent work on FSS focuses on modeling one-to-one correspondence between support and query
pixels to create correlation scores~\cite{hsnet, democratic, pgnet}. Another line of research is
learning {\em class prototypes} to be used in deciding whether each pixel belongs to the object or
the background \cite{mininglatentclasses, adaptiveprtotypelearning, SCL, pfenet,
canet, panet, amp}.  To build models that are trained to generalize over few training samples, recent work
widely adopts the {\em meta learning} paradigm~\cite{hsnet, democratic, pgnet, mininglatentclasses,
adaptiveprtotypelearning, SCL, pfenet, canet, panet} . In meta learning, the main
idea is to create a series of tasks based on the training set simulating the few-shot segmentation
problem and train the meta model over these tasks. As an alternative, it has also been recently shown
that carefully designed transductive inference mechanisms combined with simple pretraining
and fine-tuning strategies can yield state-of-the-art results without involving meta learning
procedures~\cite{repri}.

Even though the few-shot setting enables the models to make predictions for the new classes of
objects with just a few annotated images, these models still need to be trained on a large number of
base training classes to achieve strong few-shot generalization. Thus, just like the
fully-supervised approaches, these models still rely on large amounts of training examples for
training the meta model (or pretraining the base model). As a result, even the few-shot learning
approaches end up being practically limited due to the difficulties of collecting rich segmentation
training data sets. Since a meta (base) few-shot segmentation model is effectively a deep
segmentation architecture, the limitations in base training data is likely to affect the
generalization abilities of the learned meta model to a great extent.

To study ways to reduce the annotation dependency, we explore the problem of self-supervised
learning of few-shot segmentation models. The goal is to learn a class-agnostic meta model without using any segmentation annotations during training. Once the self-supervised meta
model is learned, the segmentation model of a novel class is obtained on-the-fly using a single 
sample, just as in standard FSS. Since meta model learning requires no manual annotations, such an
approach, in theory, can allow leveraging arbitrarily large-scale and rich unlabelled image
collections. With a similar motivation, a self-supervised FSS approach for medical images is first proposed by {\cite{self-supervised_fss}}. We generalize the problem to natural images with complex scenes.

In order to by-pass the dependency on supervised segmentation data for the base classes and realize
self-supervised meta-learning for FSS, we propose a saliency-based scheme for creating a training
pair from each training image. More specifically, we create episodic training tasks by (i)
estimating an unsupervised saliency mask at each image, (ii) splitting the mask stochastically into
two parts, and (iii) treating the splitted mask pairs, after image augmentations, as support and
query pairs for training the FSS model. The combination of mask splitting and strong
augmentations effectively create challenging one-shot segmentation tasks.
Figure~\ref{fig:intro} illustrates the proposed approach. 

There has been significant progress in self-supervised learning over the past few
years~\cite{selfsuprotation,selfsupsimclr,selfsupsimclr2,selfsupjigsaw,selfsupefros,selfsuppainting,selfsuporder1,selfsuporder2}.
In particular, contemporary methods obtain representations that are competitive with their supervised
counterparts when applied to recognition tasks such as image classification, object detection or semantic
segmentation. In contrast, in this paper we learn the few-shot segmentation model itself in a
self-supervised manner, and to the best of our knowledge, ours is the first one in this direction for natural images. 

We conduct extensive experiments on the widely used FSS benchmarks based on the MS-COCO~\cite{coco}
and PASCAL~\cite{pascaldataset}. We also present a detailed ablative study, where we experimentally
analyze the model over a variety of supervision settings and model component configurations.

In the remainder of the paper, we first present an overview of the recent developments on 
(few-shot) semantic segmentation and self supervised learning. We then explain
our proposed method, which consists of the problem setup, the novel self-supervised training procedure and 
the architecture of our FSS network. Finally, we present a detailed experimental analysis of the proposed framework and conclude the paper.

\section{Related work}

\mypar{Fully-supervised \& few-shot semantic segmentation} In fully-supervised semantic
segmentation, a central challenge is obtaining high-resolution segmentation results by efficiently
modeling both contextual and local information.  To incorporate the contextual information
efficiently, \cite{deeplab,dilated} introduce {\em dilated convolution}, which allows the
enlargement of the receptive field of a convolutional kernel, without increasing the number of
trainable parameters.  To tackle the same problem, pooling mechanisms, such as {\em global average
pooling}~\cite{parsenet}, {\em pyramid pooling module}~\cite{pspnet}, and {\em atrous spatial
pyramid pooling}~\cite{deeplab}, also offer powerful modeling tools. Encoder-decoder like
architectures are similarly widely used to design efficient and effective semantic segmentation
networks, \eg \cite{fcn,unet,laplacian,refinenet,deeplabv3}. Attention mechanisms are also used to improve 
long range interactions across regions, \eg \cite{psanet,contextencoding,cooccur,danet,ccnet}.

For few-shot segmentation, the pioneering work of \cite{oslsm} proposes a two-branch solution where
the conditioning branch predicts the task parameters using the support set, and then these
parameters guide the segmentation branch in predicting pixel-wise labels. Follow-up works can be
categorized as prototype-based, graph-based and meta-learning free approaches. In prototype-based
approaches, the aim is to obtain features from support examples that summarize classes with pooling
and those support features are typically matched with query features via a distance
metric~\cite{dong2018,panet,ppm,canet,rakelly2018}. In order to create stronger connections between
the support and query features, graph-based approaches \cite{pgnet} \cite{democratic} \cite{scaleawaregnn} are also used.
Since most prototype-based approaches rely on global average pooling, the aim of graph based
approaches is to establish more local-to-local connections between the support and query features. Another approach with the same objective is \cite{cyclicmemory} where a memory network is trained with different query feature resolutions.

In contrast to these meta-learning approaches, \cite{repri} first trains a segmentation
representation over the base classes using supervised training, and then uses transductively
regularized fine-tuning to obtain task-specific models.
 
\mypar{Self-supervised learning} Self-supervised learning focuses on extracting supervision from the
structure of data. 

Recently, it has been shown that high-level semantic visual representations can be successfully
learned by using self-supervision and directly used in downstream tasks such as
classification, detection and segmentation tasks
\cite{selfsuprotation,selfsupsimclr,selfsupsimclr2}. Some pretext tasks for self-supervision involve
solving jigsaw puzzles \cite{selfsupjigsaw,selfsupefros}, rotation prediction
\cite{selfsuprotation}, emptied pixel prediction \cite{selfsuppainting}, and order prediction
\cite{selfsuporder1,selfsuporder2}. 

Self-supervision based training signals have recently been used to improve FSS models. In particular,
\cite{selfsuptuning} proposes to get more refined
support features by using self-supervision on support images through inner gradient optimization.
\cite{SCL} similarly aims to obtain improved
support features with self-supervision over the support masks.
Our work fundamentally differs from both of these approaches as we aim to formulate 
a purely self-supervised approach to learn the meta-model in an unsupervised manner,
as opposed to defining an auxiliary self-supervised training loss to improve the model in a traditional
FSS setting.

A related problem is {\em unsupervised semantic segmentation}, where the goal is to cluster pixels
across images into semantic semantic groups. This problem has recently been tackled using
end-to-end \cite{iic,autounsupseg}, and two-staged approaches \cite{maskcontrast}.  Our
work fundamentally differs in terms of both the problem definition and the overall approach.  First,
instead of unsupervisedly learning a segmentation model of a fixed number of classes, we aim to
learn a class-agnostic meta model that can synthesize semantic segmentation of an arbitrary novel
class based on a single training example, in an unsupervised manner. Second, as opposed to
the clustering based approaches in unsupervised semantic segmentation, our approach relies on episodic
meta-learning over self-supervisedly generated pseudo-groundtruths.

Finally, we should note that few works have recently explored self-supervised meta-learning in a
few-shot classification context: \cite{hsu2019unsupervised} and \cite{khodadadeh2019unsupervised}
proposes clustering (and augmentation) based task creation schemes for learning few-shot
classification models.  To the best of our knowledge, ours is the first work to propose and study
self-supervised learning of a few-shot segmentation model.

\section{Method}

In this section, we first formally define the self-supervised few-shot segmentation problem, and
then the proposed training methodology. Finally, we define our network architecture realizing the proposed 
approach.

\subsection{Preliminaries and problem definition}

\def\Dtr{D_\text{train}}
\def\Dte{D_\text{test}}
\def\Ctr{C_\text{train}}
\def\Cte{C_\text{test}}
\def\x{x}
\def\m{m}
\def\y{y}

\mypar{Traditional few-shot segmentation} The ultimate goal of few-shot segmentation is to obtain 
a meta model that can yield an accurate segmentation model of a novel class,
given just one or few samples for the novel class. In the standard FSS scenario, the FSS model itself is
meta-learned (or pretrained) over a supervised training set $\Dtr$ over classes $\Ctr$, and
evaluated over a test set $\Dte$ over classes $\Cte$. Since the goal is to learn an FSS model
that generalizes well to novel classes, $\Dtr$ and $\Dte$ consists of distinct classes, \ie
$C_{test} \cap C_{train} = \emptyset$. In these data sets, each example corresponds to a triplet
$(\x,\m,\y)$ where $\x$ is the image, $\m$ is the groundtruth binary mask and $\y$ is the class label
corresponding to the mask.

\mypar{Episodic training} Meta-learning of an FSS model is typically formulated in terms of {\em
episodic training}. In episodic training, the meta model is trained over a series of training
batches consisting of support set $S$ and query set $Q$ examples, sampled from $\Dtr$. On each query
example with some class $y$, the corresponding binary mask is estimated using the meta model
according to the support samples provided for the same class.  The meta model is iteratively updated
over the episodes to minimize a semantic segmentation loss, \eg pixel-wise cross entropy loss,
evaluated on the query samples. Once the training is over, the model is tested by sampling random
episodes from $D_{test}$, where the groundtruth masks for the support samples of novel classes are
provided as one(few)-shot guidance to the meta model and those of the query samples are used only for evaluating the resulting FSS outputs on the corresponding queries.

\mypar{Self-supervised few-shot segmentation} We now define the self-supervised few-shot
segmentation problem, using the same notation as above. Similar to the standard FSS problem, in
self-supervised FSS, we are given datasets $D_{train}$ and $D_{test}$.  
Unlike standard FSS, however, here $D_{train}$ consists of only unlabeled
images, with no masks or class labels. Therefore, it is not immediately clear how to define an
episodic training procedure, as we can neither provide support samples with class-specific masks,
nor sample support and query image pairs from $\Dtr$ such that the support and query images are
known to belong to the same class.  Once the model is trained, we use the same evaluation
protocol of standard FSS to evaluate the learned meta model on large number of few-shot segmentation tasks. 

In this work, we intentionally focus on the one-shot segmentation problem to study the problem 
isolated from the orthogonal concerns regarding the fusion of guidance provided by
multiple support samples during evaluation.

\subsection{Proposed approach}
\label{subsection:training}

\begin{figure}[t]
    \includegraphics[width=.20\linewidth]{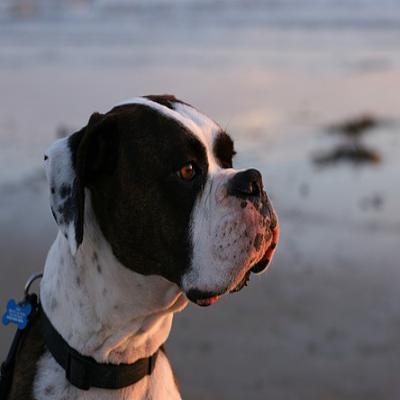}\hfill
    \includegraphics[width=.20\linewidth]{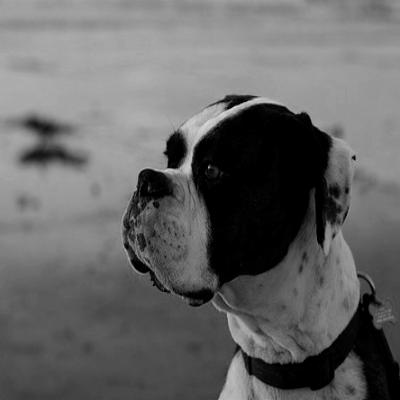}\hfill
    \includegraphics[width=.20\linewidth]{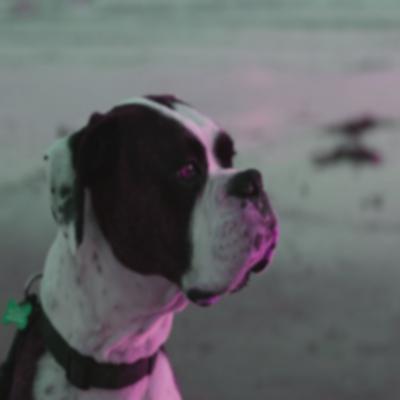}\hfill
    \includegraphics[width=.20\linewidth]{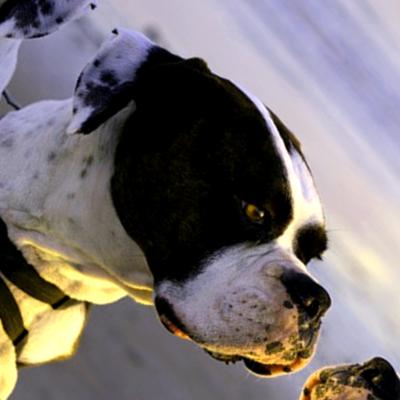}\hfill
    \includegraphics[width=.20\linewidth]{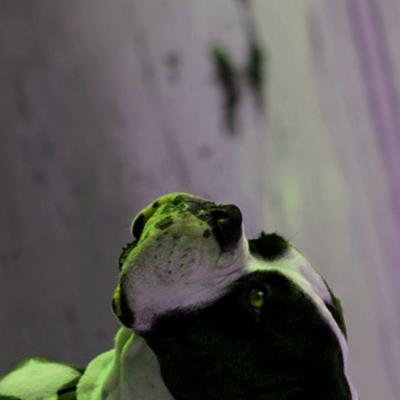}
    \\[\smallskipamount]
    \includegraphics[width=.20\linewidth]{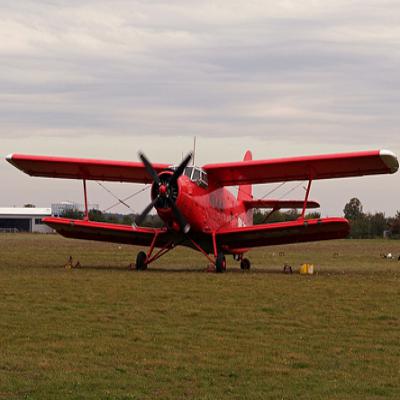}\hfill
    \includegraphics[width=.20\linewidth]{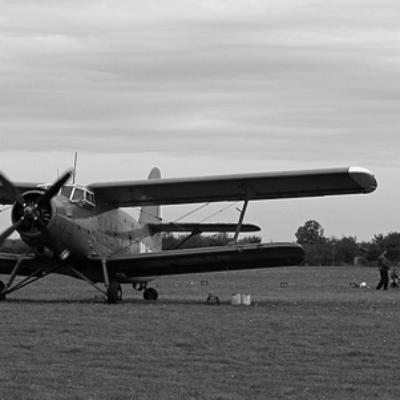}\hfill
    \includegraphics[width=.20\linewidth]{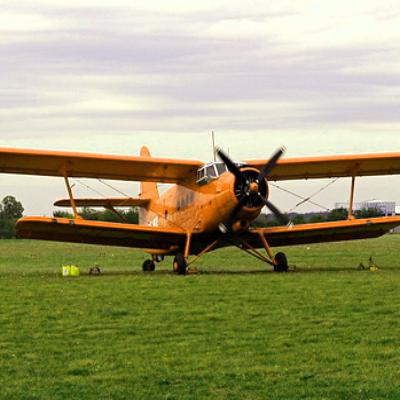}\hfill
    \includegraphics[width=.20\linewidth]{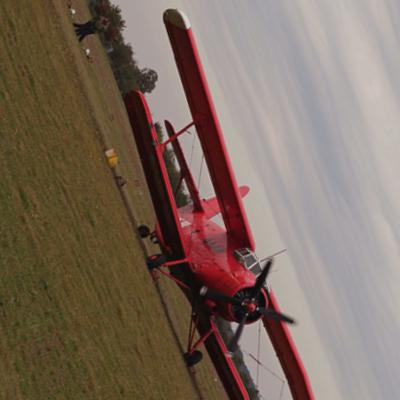}\hfill
    \includegraphics[width=.20\linewidth]{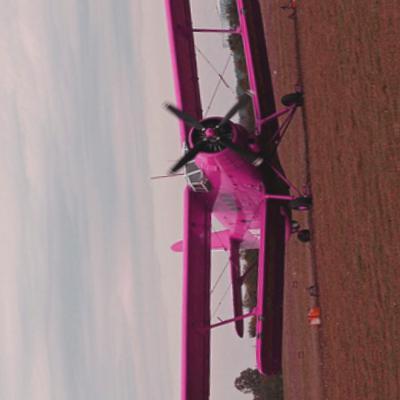}
    \\[\smallskipamount]
    \includegraphics[width=.20\linewidth]{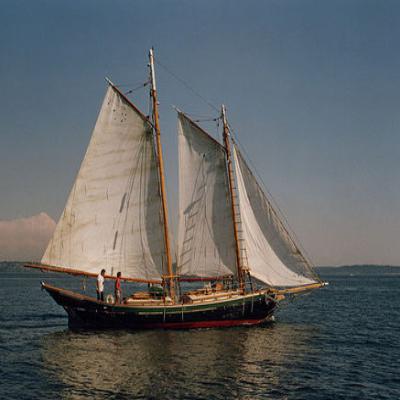}\hfill
    \includegraphics[width=.20\linewidth]{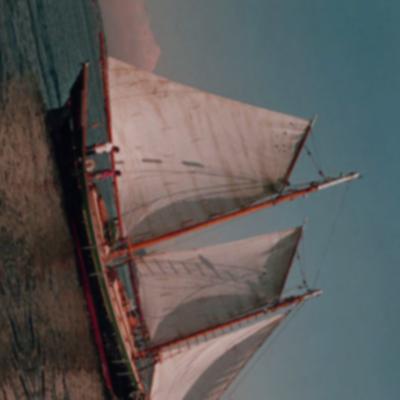}\hfill
    \includegraphics[width=.20\linewidth]{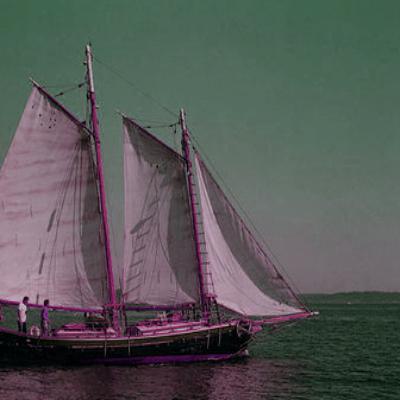}\hfill
    \includegraphics[width=.20\linewidth]{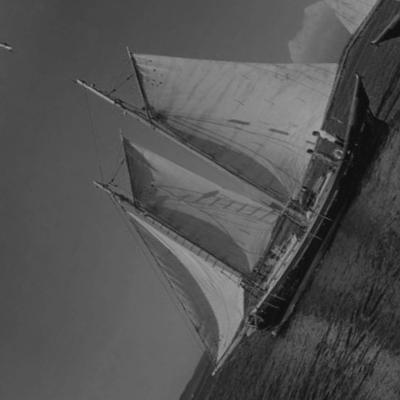}\hfill
    \includegraphics[width=.20\linewidth]{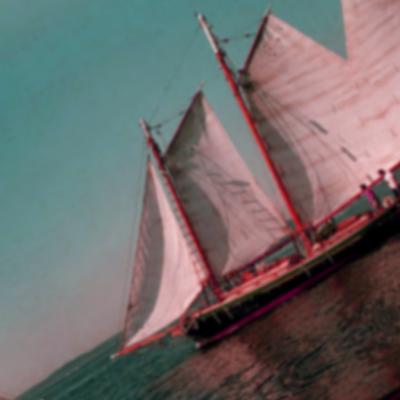}
    \\[\smallskipamount]
    \includegraphics[width=.20\linewidth]{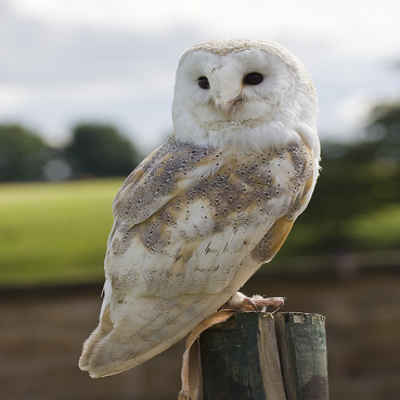}\hfill
    \includegraphics[width=.20\linewidth]{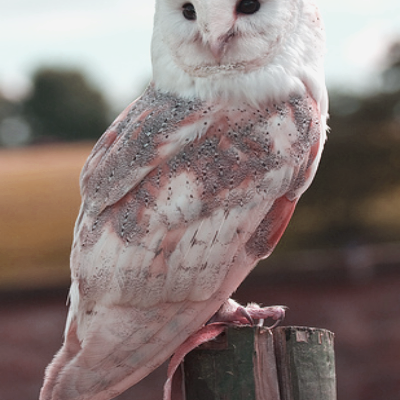}\hfill
    \includegraphics[width=.20\linewidth]{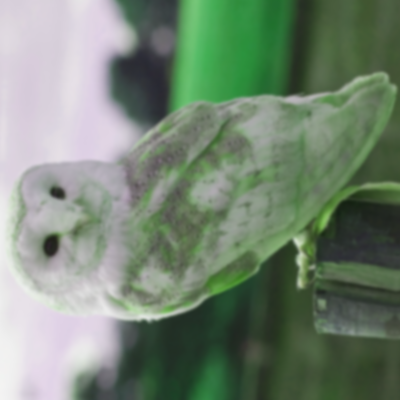}\hfill
    \includegraphics[width=.20\linewidth]{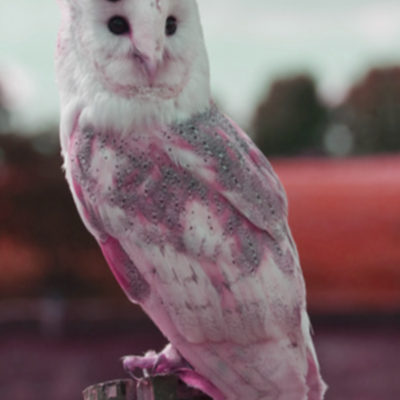}\hfill
    \includegraphics[width=.20\linewidth]{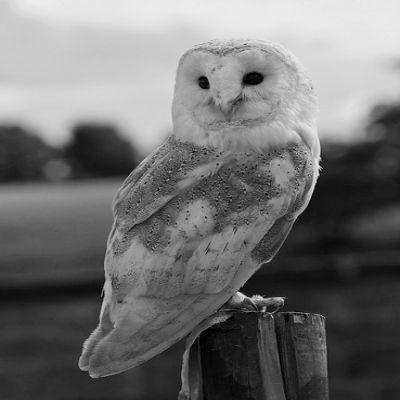}
    \caption{Augmentation samples. The first column show the original images and the remaining four columns contain images with different augmentations.}\label{fig:augmentations}
    \vspace{-4mm}
\end{figure}

\mypar{Saliency-driven self-supervised training} As explained above, in the proposed self-supervised
FSS problem, the training set lacks masks and labels. To address the first problem, \ie the lack of
groundtruth masks, we propose to use unsupervised saliency to define pseudo groundtruth masks. For
this purpose, we adapt the saliency mask estimation approach used in \cite{maskcontrast}: we train
an unsupervised saliency model on the MSRA dataset~\cite{878f4ef5fd274e93b8b3f9f46b1c3872} using the
DeepUSPS approach~\cite{DeepUSPS}. We then train a BAS-NET model~\cite{BasNet} from scratch using
the saliency estimations given by the unsupervised saliency model. Then BAS-NET model trained on
pseudo-masks is used to obtain object mask proposals from Pascal and COCO datasets.

While the use of unsupervised saliency estimates provides an alternative to manual groundtruth
masks, it still does not address the second main problem, \ie the lack of class labels in $\Dtr$,
which are also required to form same class support and query pairs.  To remedy this problem, we
propose to create support and query pairs from each individual image, by adapting contemporary self-supervised
representation learning practices and leveraging the spatial
nature of the segmentation task.

More specifically, to define an episode from a single image, we need to create distinct support and
query samples. To achieve this, we propose to first apply a set of random augmentations twice to each training image. To this
end, we adapt the augmentations used in the SimCLR~\cite{selfsupsimclr}, and utilize the 
\texttt{grayscale, color jitter, horizontal flip, vertical flip, rotate} and \texttt{random resize}
augmentations. We emphasize again that while the SimCLR method aims to learn a global image representation based on contrastive learning over augmented image patches,
our goal is to episodically meta-learn a one-shot segmentation model by
forming
support and query pairs that differ significant enough from each other. 
As can be observed in Figure~\ref{fig:augmentations}, the utilized augmentations are able to produce a wide range of 
variants of a single image.

\begin{figure}[t]
    
    \centering
     \begin{subfigure}[b]{0.8\linewidth}
         \centering
         \includegraphics[width=.25\linewidth]{images/augmentation_images/2008_004752/original.jpg}\hfill
        \includegraphics[width=.25\linewidth]{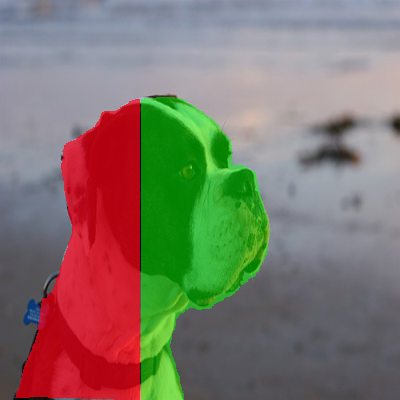}\hfill
        \includegraphics[width=.25\linewidth]{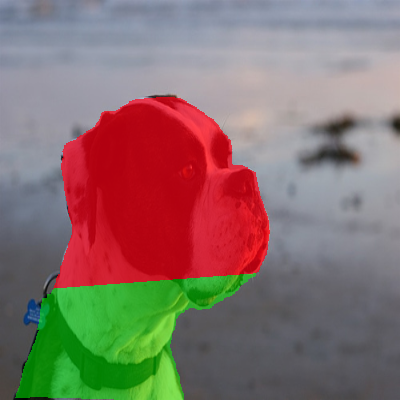}\hfill
        \includegraphics[width=.25\linewidth]{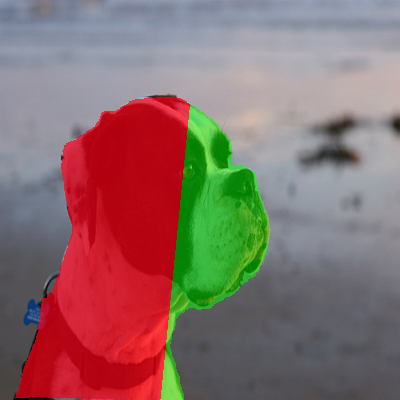}
     \end{subfigure}\\[\smallskipamount]
    \centering
     \begin{subfigure}[b]{0.8\linewidth}
         \centering
         \includegraphics[width=.25\linewidth]{images/augmentation_images/2010_001413/original.jpg}\hfill
        \includegraphics[width=.25\linewidth]{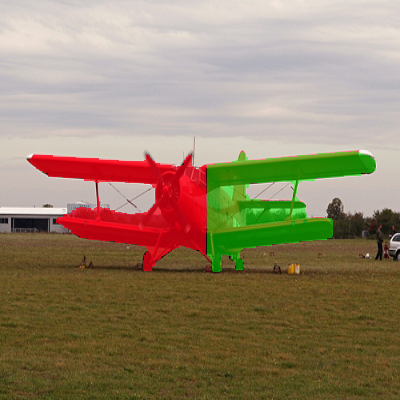}\hfill
        \includegraphics[width=.25\linewidth]{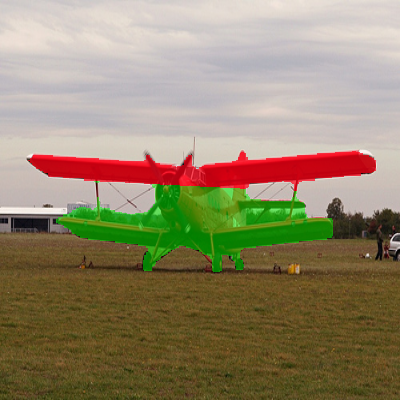}\hfill
        \includegraphics[width=.25\linewidth]{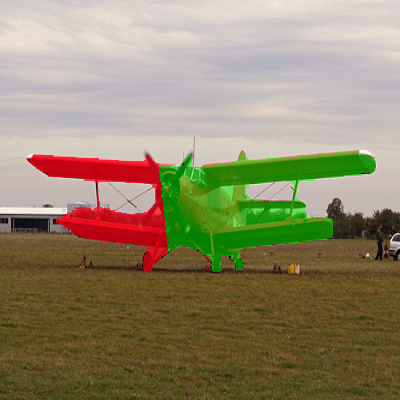}
     \end{subfigure}\\[\smallskipamount]
    \centering
     \begin{subfigure}[b]{0.8\linewidth}
         \centering
         \includegraphics[width=.25\linewidth]{images/augmentation_images/2010_000748/original.jpg}\hfill
        \includegraphics[width=.25\linewidth]{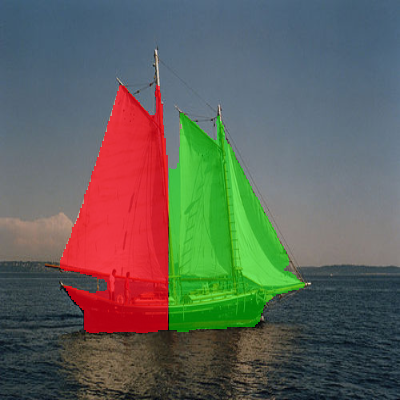}\hfill
        \includegraphics[width=.25\linewidth]{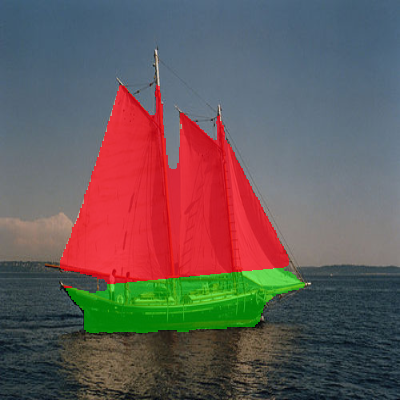}\hfill
        \includegraphics[width=.25\linewidth]{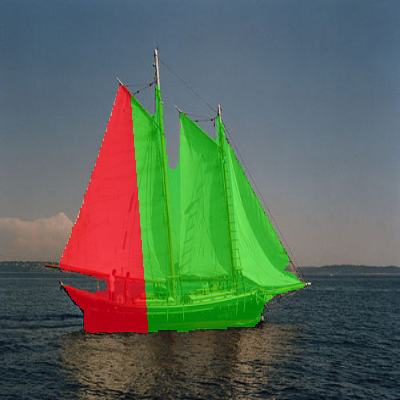}
     \end{subfigure}\\[\smallskipamount]
    \centering
     \begin{subfigure}[b]{0.8\linewidth}
         \centering
         \includegraphics[width=.25\linewidth]{images/augmentation_images/2009_003685/original.png}\hfill
        \includegraphics[width=.25\linewidth]{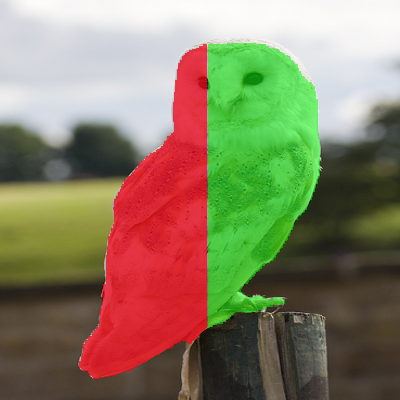}\hfill
        \includegraphics[width=.25\linewidth]{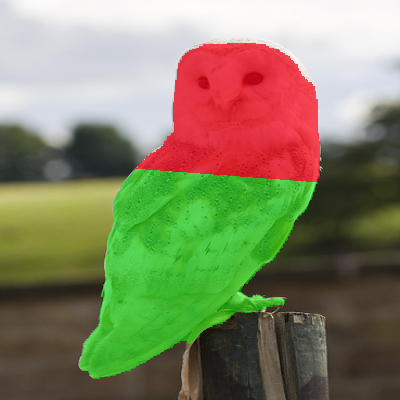}\hfill
        \includegraphics[width=.25\linewidth]{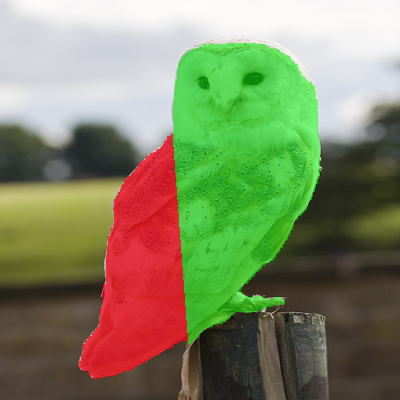}
     \end{subfigure}
    
    \caption{Splitting samples. The images on the first column are original input images and the remaining three columns contain images that demonstrate Vsplit, Hsplit with slope and Vsplit with slope splits overlaid, respectively.\label{fig:splits}}
    \vspace{-4mm}
\end{figure}

While augmentations are effectively used in self-supervised learning of image-wide representations,
here we target a more structured and arguably detailed task, \ie to learn the few-shot learning
meta-model for producing pixelwise predictions. Hence, we seek for more powerful ways to
construct training pairs. For this purpose, we propose a method that we call {\em MaskSplit} in
general.  The main idea is rooted in the observation that different parts of a single object
visually differ significantly. Based on this insight, we propose to split each saliency
mask approximately in half, randomly treat one side as the {\em support} foreground mask and the other side 
as the {\em query} foreground mask. By using these masks over the separately augmented versions of an image,
we effectively construct support and query pairs, and by using these pairs episodic training can be formed with no 
groundtruth masks or class labels (Figure~\ref{fig:intro}).

We consider three main variants of the MaskSplit framework: vertical splitting ({\em Vsplit}),
horizontal splitting ({\em Hsplit}), and their combination ({\em MixedSplit}). To avoid potential
unwanted biases caused by always using axis-aligned mask splits, we split masks along an oriented
line. We refer to the basic axis-aligned versions of these schemes as splitting {\em without slope}.
We present visual examples corresponding to Vsplit without slope, Hsplit with slope and Vsplit with slope in
Figure~\ref{fig:splits}, and provide more technical details in the following paragraphs.

\begin{figure*}[ht]
\centering
  \includegraphics[width=0.8\linewidth]{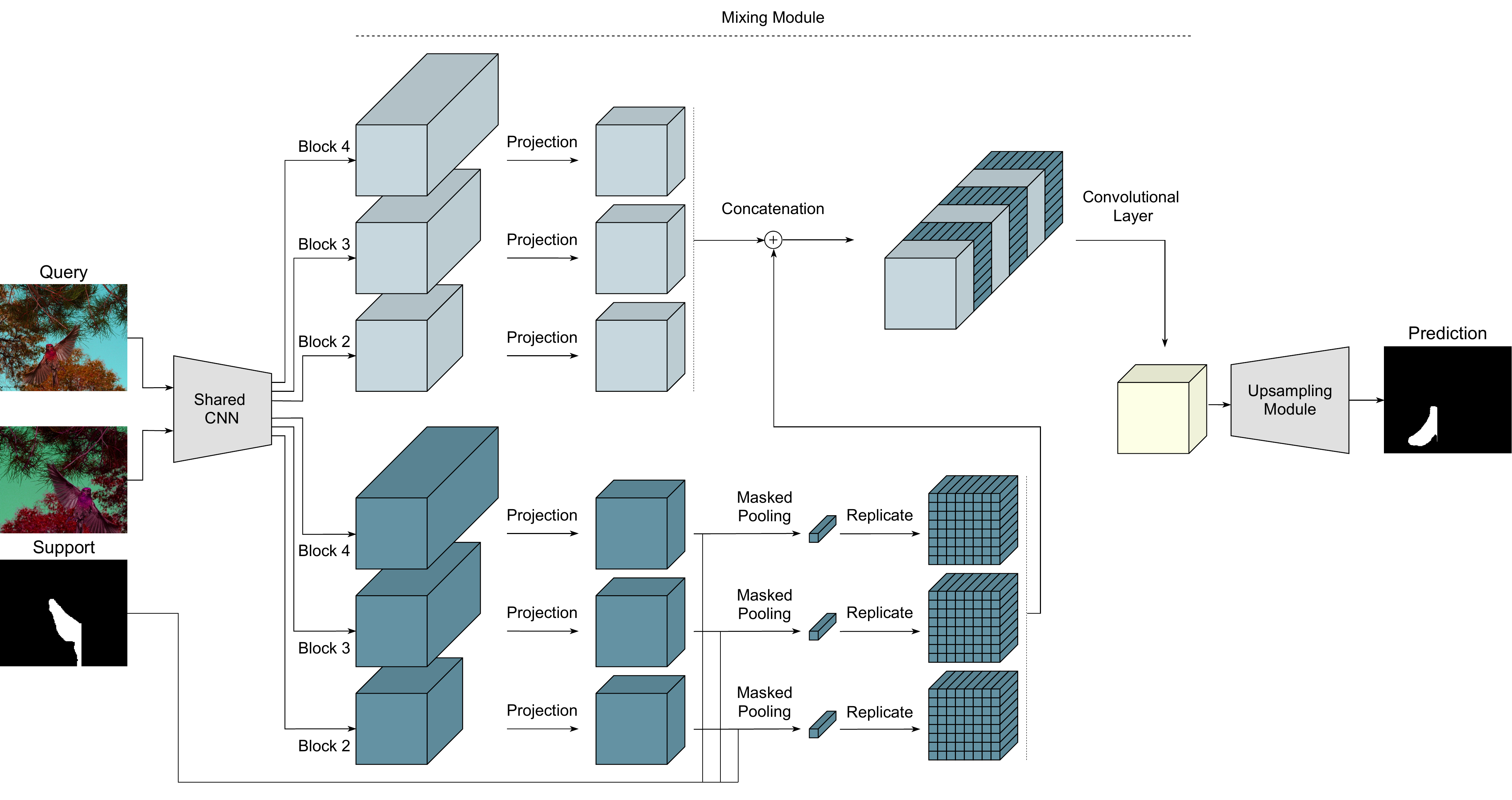}
  \caption{Architecture of our prototype based network, trained with the proposed self-supervised meta-learning approach.}
  \label{fig:mainarch}
  \vspace{-4mm}
\end{figure*}

\mypar{Vsplit} To divide images in half using Vsplit, we find the line $l_0$ parallel to
y-axis of the image such that $l_0$ divides the foreground pixels in half. We then assign right side
to be the support's foreground and the left side to be the query's foreground. To
obtain more and different combinations of this splitting procedure we propose {\em alternating} vertical
split with slope.  Here, alternating means that  randomly assigning left or right side of saliency
mask to support and query instead of assigning them to same-side in every episode. In order to split
the foreground with slope, we shift the line by a number of pixels on both top of the image and the
bottom of the image. The shift operation is done in opposite directions for the top and bottom
intersection of the image with the line. The default shift range is $(-40,40)$ in our experiments.

\mypar{Hsplit and MixedSplit} We also perform experiments with horizontal splitting with slope, where we
find the line $l_0$ parallel to x-axis instead of y-axis. The remaining procedure is the same as in
Vsplit. In MixedSplit, we apply Vsplit to the fifty percent of the episodes and apply
Hsplit to the rest. 

An important implementation detail is the definition of areas over which the loss function is
evaluated. More specifically, it shall be observed that part of the non-foreground region in a
query image corresponds to the support pixels. Therefore, a naive implementation of the
query loss might enforce negative predictions over those support pixels on the query image.
To avoid this problem, we ignore the query pixels corresponding to the support area by not
calculating the loss function in those background pixels.

\subsection{Network architecture}
\label{subsection:arch}

We define a simple prototype-based model inspired from the FSS architecture of \cite{canet} to explore different self-supervised strategies. Our network consists of three parts, a \textbf{backbone network} to extract features from query and support images, a \textbf{mixing module} where support and query features are mixed together, and an \textbf{upsampling network} where the mixed features are used to predict the segmentation mask. Figure~\ref{fig:mainarch} illustrates the structure of our network.

\mypar{Backbone} We use ResNet-101~\cite{resnet101} as our backbone to extract features
from query and support images. ResNet-101 consists of 4 main blocks that are composed of several
convolutional layers. It has been shown that in the first layers of a ConvNet, the model focuses on
low-level geometrical features, whereas final layers focus on higher-level semantic
features~\cite{convnet-visu}. In our work, we want to take advantage of both geometric and semantic
cues by using all blocks, except $block1$.  Normally, after each block, the spatial resolution is
decreased by strided convolutions. To preserve the spatial resolution we use dilated
convolutions~\cite{deeplab} instead of strided convolutions after $block2$ so that all feature
maps after block2 has a fixed size of 1/8 of the original image.  To be able to compare to the
existing work on FSS, we use a backbone model pretrained on the ImageNet dataset~\cite{imagenet}.

\mypar{Mixing module} To mix different levels of cues we propose a feature mixing module to both mix semantic and geometric features and shape the query features for the prediction using support prototype. The mixing module consists of two sub-modules: projection module to project channel dimensions of different level of features onto a fixed channel size and shared convolution module to mix support-aware query features. There are three projection layers for each block, respectively $block2,block3,block4$. 
After both query and support features are projected onto a fixed channel dimension, for each different feature level, the query features are concatenated with their corresponding global-average pooled support prototypes. Then all these different levels are concatenated into a feature map and passed onto a shared convolution layer.
The output of this module is used by the upsampling module to make predictions.

\mypar{Upsampling Module} We use a simple upsampling module to decode support-aware query features
into a segmentation mask. It consists of 4 blocks, the first 3 blocks composed of 2 convolutional
layers and a bilinear upsampling operation. The last block is composed of one $3{\times}3$ convolutional layer
followed by a $1\times1$ convolutional layer.
\vspace{-3mm}

\section{Experiments}

\begin{table*}[ht]
\begin{center}
\resizebox{.75\linewidth}{!}{
\begin{tabular}{ c c c c c c| c c c c c }
\toprule
 \texttt{aug} & \texttt{slope} & \texttt{alternate} & \texttt{Vsplit} & \texttt{Hsplit} & \texttt{prob} & $5^0$ & $5^1$ & $5^2$ & $5^3$ & mean \\
 
\hline
\checkmark &  &  & & & 0.0 & 51.5 & 51.1 & 52.1 & 40.0 & 48.6  \\

 & \checkmark & \checkmark & \checkmark & & 1.0 & 50.7 & 50.1 & 49.7 & 40.8 & 47.8  \\

\checkmark & & \checkmark & \checkmark &  & 1.0  & 53.2 & 54.5 & 54.0 & 42.4 & 51.0 \\

\checkmark & \checkmark & \checkmark &  & \checkmark & 1.0 & 54.9 & 55.9 & 54.7 & 43.9 & 52.3 \\

\checkmark & \checkmark & \checkmark & \checkmark & \checkmark & 1.0 & 54.5 & 55.5 & 52.5 & 42.7 & 51.3  \\

 \checkmark & \checkmark &  & \checkmark & & 1.0  & 53.7 & 57.1 & 55.4 & 44.7 & 52.7  \\

\hline
\checkmark & \checkmark & \checkmark & \checkmark & &  0.3  & \textbf{54.4} & 54.2 & \textbf{55.4} & 43.7 & 51.9  \\
\checkmark & \checkmark & \checkmark & \checkmark & & 1.0 & 54.1 & \textbf{57.1} & 54.9 & \textbf{46.1} &  \textbf{53.0}   \\
\bottomrule
\end{tabular}}
\end{center}
\vspace{-3mm}
\caption{Ablation study of our proposed MaskSplit framework. We report the mIoU scores achieved by different variants of our model on the PASCAL dataset. \texttt{Prob} stands for the probability of applying the selected image splitting method. 
}
\label{table:vsplitvariants}
\vspace{-3mm}
\end{table*}

\mypar{Datasets}
To evaluate our approach, we use two standard few-shot semantic segmentation datasets: PASCAL-$5^i$ and COCO-${20}^i$. Pascal-$5^i$ dataset is proposed in OSLSM \cite{oslsm} and is based on a combination of PASCAL VOC 2012 \cite{pascaldataset}, and the extra annotations from \cite{sbd}. This dataset contains 20 classes that are evenly divided into four folds. In the literature, for each fold $i$, the other three folds are used to train the model, and fold $i$ is used as the target fold.  However, in the case of unsupervised meta learning, it is not necessary to divide the images into groups based on their class information. We can utilize the entire training data without any class information to learn our self-supervised representation. Since the training relies only on unsupervised extraction of saliency maps, this enables us to use the images from all folds to train our model. We evaluate our model with and without fold-specific training.

The second dataset, COCO-${20}^i$, is proposed by FWB \cite{fwb} and is based on COCO dataset \cite{coco}. This dataset consists of 80 classes, divided evenly into four folds. Our results for this dataset are obtained by unsupervised training on images from all folds. For both datasets, we use mean intersection-over-union (mIoU) and report one-shot segmentation results in all experiments.

\mypar{Implementation Details}
For the backbone network we use ResNet-101. We extract features from $block2, block3, block4$, and then these features are passed onto the mixing module to mix query and support features at different levels. 
Both query and support images are resized to a fixed spatial size of $400{\times}400$ and features extracted from the ResNet-101 backbone have a fixed size of $50\times50$. The overall architecture is implemented in PyTorch and PyTorch-Lightning. We use Adam optimizer with a learning rate of $10^{-4}$ and weight decay $10^{-5}$ with pixel-wise cross-entropy loss. All layers of ResNet-101 are kept frozen. We train each model for 100 epochs on PASCAL dataset and 20 epochs on COCO dataset with a batch size of 16, on a single Nvidia V100 GPU. To test the models, we use 5 runs with 2500 tasks each.

\subsection{Ablation Study}

We first conduct extensive ablative studies to understand the effects of major components of our approach such as \texttt{Alternate}, \texttt{Slope}, \texttt{Vsplit} and \texttt{Aug}, on the Pascal-$5^i$ dataset. We also experiment with different splitting techniques \texttt{Hsplit}, \texttt{MixedSplit} and also with no split. Finally, we show the results of different supervision levels. 

\mypar{MaskSplit variants} Here, we look at the effect of different components of our approach. These components are augmentations (denoted with \texttt{aug}), slope (denoted with \texttt{slope}), alternating support and query sides (denoted with \texttt{alternate}), probability of applying the split (denoted with \texttt{prob}), the type of split (vertical split \texttt{Vsplit} or horizontal split \texttt{Hsplit}).
When splits are are applied with 0.3 probability, we apply MaskSplit 30 percent of the time for the given configuration, and for the remaining, we only apply augmentations and use the saliency mask as both support's and query's foreground. 

The results are summarized in Table \ref{table:vsplitvariants}. The first row of Table \ref{table:vsplitvariants} corresponds to the case when SimCLR \cite{selfsupsimclr} augmentations are used together with our base prototype network. This self-supervised version, makes use of only the SimCLR augmentations instead of MaskSplit based training. This yields a mIoU of $48.6$ on average. When we do not use any augmentations, but use the Vsplit, we see that the average mIOU score is $47.8$. When Vsplit is used together with the augmentations (using \texttt{aug}, \texttt{slope}, \texttt{alternate} components), we achieve a average mIoU of $53.0$. This proves that augmentations are necessary to create visually different query and support examples, and including different augmentations within the self-supervised learning procedure enhances the training quite significantly. This significant performance increase from $48.6$ to $53.0$ also indicates that proposed splitting mechanism is effective for the self-supervised segmentation task.

We also observe in Table \ref{table:vsplitvariants} that removing slope option causes the mIoU to drop by $2$ points, which means that slope improves the possibility of creating variety of segments in episodes. Making Vsplit probabilistic reduces mIoU approximately one percent. mIoU scores show that applying Vsplit only thirty percent of the time is also effective, if not as effective as applying it all the time. 

Lastly, removing the \texttt{alternate} option yields only 0.3 point reduction in average mIoU. Here, applying augmentations probably render the sides of the splits to be different enough at each episode. Therefore, not alternating between sides does not seem to have a major effect; nevertheless, it still is a useful part of the procedure.

\mypar{Different Split Techniques}
 Table \ref{table:vsplitvariants} also presents the results with using different split techniques. When no split is used, there is a decrease of almost 4.5 percent in mIoU. This result highlights that proposed splitting procedure adds significant recognition power to the self-supervised process. The comparison between results of \texttt{Hsplit} and \texttt{Vsplit} show that the positioning of objects in the images are most appropriate to be used with \texttt{Vsplit}. 
 For the rest of the experiments, we use MaskSplit with \texttt{Vsplit},  \texttt{aug,slope,alternate} options, which achieves the best performance in these ablation studies.

\mypar{Different Supervision Levels}
We conduct further experiments to see the effect of using different supervision levels and different amount of training data. 
The first two rows of Table \ref{table:supervision} compare the results of MaskSplit using regular fold-based training; \ie using groundtruth masks during training vs masks obtained in the unsupervised fashion using saliency. Since the number of training images is relatively low in fold-based training, using groundtruth masks only is limiting the context information learned for the self-supervised learner. 
On the contrary, saliency maps could also include objects that are not in the groundtruth masks. As a result, self-supervised learning appears to be positively affected by the increased variety.

For the supervised counterpart, we train our base prototype network using traditional fold-based few-shot semantic segmentation setting. We observe that this supervised training of the model yields less superior results compared to the self-supervised version that is trained over the entire set of training images (52.6 mIoU vs 53.0 mIoU). The self-supervised nature of training enables us to use all the training images, and this leads to a better learning of segmentation in an unsupervised way.

The last row of Table \ref{table:supervision} shows the oracle results of our model, which uses ground truth segmentation maps instead of masks extracted by unsupervised saliency model. This gives us a upper limit on the performance of our proposed framework. The results demonstrate that when we use groundtruth masks instead of saliency, 57.3 percent mIoU score can be achieved.

\begin{table}
\begin{center}
\resizebox{\linewidth}{!}{
\begin{tabular}[h]{ l c c c c c c c }
\toprule
supervision & mask source & train & $5^0$ & $5^1$ & $5^2$ & $5^3$ & avg  \\
\hline
Self-sup. & groundtruth      & fold      & 49.4  & 52.8  & 40.6  & 37.6  & 45.1  \\
Self-sup. & saliency          & fold      & 51.5  & 55.2  & 52.5  & 44.4  & 50.9 \\
Supervised      & groundtruth      & fold      & 54.9  & 65.4  & 47.9  & 42.2  & 52.6 \\
Self-sup. & saliency          & all       & 54.1  & 57.1  & 54.9  & 46.1  & 53.0   \\
Self-sup. & groundtruth      & all       & 59.0  & 59.0  & 62.5  & 49.0  & 57.3  \\
\bottomrule
\end{tabular}
}
\end{center}
\vspace{-4mm}
\caption{Comparison of different supervision levels and effect of using fold-based training vs all train set training using the PASCAL dataset. For supervised, we use our base prototype model  trained in the standard supervised few-shot setting, using the groundtruth masks from the training folds. 
}
\label{table:supervision}
\vspace{-2mm}
\end{table}

\subsection{Comparison to existing work}

\begin{table}
\begin{center}
\resizebox{.9\linewidth}{!}{
\begin{tabular}{ l|c c c c c }
\hline
Method & $5^0$ & $5^1$ & $5^2$ & $5^3$ & avg  \\
\hline
\multicolumn{6}{c}{Supervised meta-learning (upper-bounds) with ResNet50} \\
\hline
PANet~\cite{panet} & 44.0 & 57.5 & 50.8 & 44.0 & 49.1  \\
PGNet~\cite{pgnet} & 56.0 & 66.9 & 50.6 & 50.4 & 56.0  \\
PFENet~\cite{pfenet} & 61.7 & 69.5 & 55.4 & 56.3 & 60.8  \\
SCL(PFENet)~\cite{SCL} & 63.0 & 70.0 & 56.5 & 57.7 & 61.8   \\
RePRI~\cite{repri} & 59.8 & 68.3 & 62.1 & 48.5 & 59.7  \\
SAGNN~\cite{scaleawaregnn} & 64.7 & 69.6 & 57.0 & 57.2 & 62.1 \\
CMN~\cite{cyclicmemory} & 64.3 & 70.0 & 57.4 & 59.4 & 62.8 \\
\hline
\multicolumn{6}{c}{Supervised meta-learning (upper-bounds) with ResNet101} \\
\hline
FWB~\cite{fwb} & 51.3 & 64.5 & 56.7 & 52.2 & 56.2  \\
PPNet~\cite{ppnet} & 52.7 & 62.8 & 57.4 & 47.7 & 55.2  \\
DAN~\cite{democratic} & 54.7 & 68.6 & 57.8 & 51.6 & 58.2  \\
MLC~\cite{mininglatentclasses} & 60.8 & 71.3 & 61.5 & 56.9 & 62.6  \\
HSNet~\cite{hsnet} & 67.3 & 72.3 & 62.0 & 63.1 & 66.2  \\
\hline
\multicolumn{6}{c}{Unsupervised approaches} \\
\hline
Saliency*~\cite{maskcontrast} & 51.5 & 49.1 & 48.1 & 39.0 &  46.9 \\
MaskContrast*~\cite{maskcontrast} & 53.6 & 50.7 & 50.7 & 39.9 &  48.7 \\
Ours & \textbf{54.1} & \textbf{57.1} & \textbf{54.8} & \textbf{46.1} &  \textbf{53.0} \\
\bottomrule
\end{tabular}}
\end{center}
\vspace{-4mm}
\caption{mIoU scores produced by our method on PASCAL dataset in comparison with two unsupervised approaches and the supervised state-of-the-art few-shot semantic segmentation models. (*) corresponds to the results acquired by adapting the methods to FSS evaluation. }
\label{table:pascalresults}
\vspace{-6mm}
\end{table}

On Table \ref{table:pascalresults} and Table \ref{table:cocoresults} we compare our approach with the state-of-the-art supervised few-shot segmentation and unsupervised semantic segmentation approaches. The supervised state-of-the-art models are trained on the usual fold-based training using groundtruth segmentation masks, whereas unsupervised models are trained using all the training images with no groundtruth. 

In the bottom part of Table \ref{table:pascalresults}, we present comparisons to the unsupervised semantic segmentation methods. First is unsupervised saliency, for which we use the version that is optimized by \cite{maskcontrast}. In evaluation, for each test episode, we take the IoU of the proposed saliency mask with the ground truth. Second unsupervised approach is the recent state-of-the-art unsupervised semantic segmentation method, namely MaskContrast~\cite{maskcontrast}. We use the publicly available model, which was initialized using MoCo v2 \cite{moco}. To obtain the few-shot results, we take the masks produced after the k-means clustering. These masks contain cluster assignments instead of ground truth classes. \cite{maskcontrast} use the Hungarian matching algorithm to match ground truth classes with cluster assignments. However, this is not directly comparable with unsupervised few-shot segmentation methods, since the Hungarian matching algorithm requires the use of ground truth masks of validation set images. Additionally, episode creation in few-shot segmentation causes a different test set distribution. In order to make all methods directly comparable, we experiment with two different settings: (i) For each episode, we compare the cluster assignments in query and support masks. If they match, we take the IoU of the mask and the ground truth. This first evaluation yields 25.9 mIoU. (ii) For each episode, we take the IoU of the mask and the ground truth without requiring cluster assignments in support and query to match. This second evaluation results in a mIoU of 48.7. We use this second favorable result for comparison. According to the results in Table \ref{table:pascalresults}, our model outperforms both baselines by at least a margin of 4\% on Pascal-$5^i$ dataset. 

The comparisons to supervised approaches in Table \ref{table:pascalresults} show that the proposed self-supervised approach performs comparable to several recent supervised approaches, and the performance gap against the state-of-the-art is arguably not drastic. 
The results highlight the potential of the self-supervised learning of FSS models. 

In Table \ref{table:cocoresults}, we also compare our model with a baseline model and state-of-the-art few-shot segmentation models on COCO-$20^i$. We do not report any result for MaskContrast, since there is not a publicly available model that is trained on COCO dataset. Our model is again able to outperform the saliency method, yet the performance increase is relatively lower. We think that this is due to the larger amount of noise in the generated saliency masks on COCO images. 

We have also tried to adapt the self-supervised super-pixel based training strategy from {\cite{self-supervised_fss}}, originally proposed for medical images. At each training image, we extract super-pixels via \cite{super-pixel-algo}, randomly select a super-pixel, and apply augmentations using our pipeline to obtain query and support regions. Despite our efforts, however, we have not been able to get a meaningful baseline in our setting.

\begin{table}
\begin{center} 
\resizebox{\linewidth}{!}{
\begin{tabular}{ l|c c c c c }
\toprule

 Method & $20^0$ & $20^1$ & $20^2$ & $20^3$ & avg  \\
\hline
\multicolumn{6}{c}{Supervised meta-learning (upper-bounds) with ResNet50} \\
\hline
 PPNet~\cite{ppnet} & 28.1 & 30.8 & 29.5 & 27.7 & 29.0  \\

PFENet~\cite{pfenet} & 36.5 & 38.6 & 34.5 & 33.8 & 25.8  \\

 RePRI~\cite{repri} & 32.0 & 38.7 & 32.7 & 33.1 & 34.1  \\

 HSNet~\cite{hsnet} & 36.3 & 43.1 & 38.7 & 38.7 & 39.2  \\
 
 CMN~\cite{cyclicmemory} & 37.9 & 44.8 & 38.7 & 35.6 & 39.3 \\  
\hline
\multicolumn{6}{c}{Supervised meta-learning (upper-bounds) with ResNet101} \\
\hline
 FWB~\cite{fwb} & 17.0 & 18.0 & 21.0 & 28.9 & 21.2  \\

DAN~\cite{democratic} & - & - & - & - & 24.4  \\

 MLC~\cite{mininglatentclasses} & 50.2 & 37.8 & 27.1 & 30.4 & 36.4  \\
 
SAGNN~\cite{scaleawaregnn} & 36.1 & 41.0 & 38.2 & 33.5 & 37.2 \\

 PFENet~\cite{pfenet} & 36.8 & 41.8 & 38.7 & 36.7 & 38.5  \\

 HSNet~\cite{hsnet} & 37.2 & 44.1 & 42.4 & 41.3 & 41.2  \\

\hline
\multicolumn{6}{c}{Unsupervised approaches} \\
\hline
Saliency*~\cite{maskcontrast} & \textbf{22.7} & 24.3 & 20.4 & 22.2 & 22.4  \\
Ours & 22.3 & \textbf{26.1} & \textbf{20.6} & \textbf{24.3} & \textbf{23.3}  \\
\bottomrule
\end{tabular}}
\end{center}
\vspace{-4mm}
\caption{Comparison of mIoU scores produced by our method and the state-of-the-art models on COCO dataset. (*) corresponds to the results acquired by adapting the methods to FSS evaluation.}
\label{table:cocoresults}
\vspace{-4mm}
\end{table}

\begin{figure}

    \includegraphics[width=.16\linewidth]{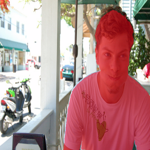}\hfill
    \includegraphics[width=.16\linewidth]{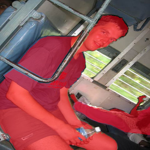}\hfill
    \includegraphics[width=.16\linewidth]{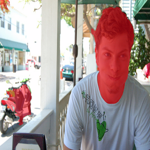}\hfill
    \includegraphics[width=.16\linewidth]{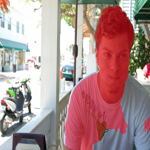}\hfill
    \includegraphics[width=.16\linewidth]{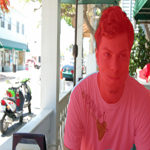}\hfill
    \includegraphics[width=.16\linewidth]{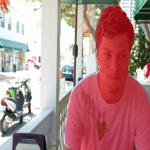}
    \\[\smallskipamount]
    \includegraphics[width=.16\linewidth]{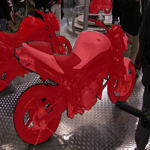}\hfill
    \includegraphics[width=.16\linewidth]{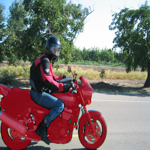}\hfill
    \includegraphics[width=.16\linewidth]{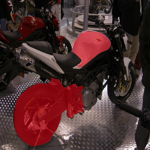}\hfill
    \includegraphics[width=.16\linewidth]{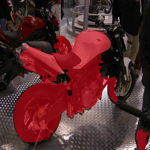}\hfill
    \includegraphics[width=.16\linewidth]{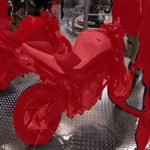}\hfill
    \includegraphics[width=.16\linewidth]{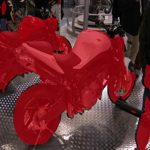}
    \\[\smallskipamount]
    \includegraphics[width=.16\linewidth]{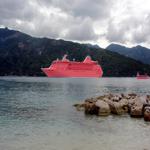}\hfill
    \includegraphics[width=.16\linewidth]{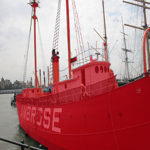}\hfill
    \includegraphics[width=.16\linewidth]{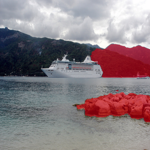}\hfill
    \includegraphics[width=.16\linewidth]{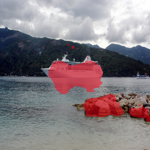}\hfill
    \includegraphics[width=.16\linewidth]{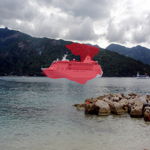}\hfill
    \includegraphics[width=.16\linewidth]{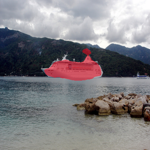}
    \caption{Qualitative results. Columns correspond to query, support, saliency map, result when trained using no augmentations, result when trained without splitting, and result of the proposed MaskSplit approach, respectively. }\label{fig:qualitativeresults}
    
\vspace{-6mm}    
\end{figure}

In Figure \ref{fig:qualitativeresults}, we present some qualitative results that demonstrate the challenges of the task and overall success of the proposed MaskSplit approach.
These results show that our method is able create accurate few-shot segmentation results, even when the saliency maps significantly differ from the groundtruth masks. In addition, the presented results also illustrate the importance of various components of our approach.

\section{Conclusion}

In this work, we define and study the problem of unsupervised few-shot semantic segmentation. Our work aims to remove the need for supervised segmentation examples for meta model training and enable utilization of arbitrarily large unsupervised image collections. We propose a novel self-supervised way to create training episodes, which is based on unsupervised saliency and augmentations. Extensive experiments show that this setting is able to achieve few-shot generalization and we obtain significant performance improvements over our baselines. We believe that our work will stimulate further study on unsupervised learning of few-shot segmentation models.

\section{Acknowledgement}
This work was supported in part by the TUBITAK Grant 119E597.

{\small
\bibliographystyle{ieee_fullname}
\bibliography{paper}

\begin{thebibliography}{10}\itemsep=-1pt

\bibitem{repri}
Malik {Boudiaf}, Hoel {Kervadec}, Ziko {Imtiaz Masud}, Pablo {Piantanida},
  Ismail {Ben Ayed}, and Jose {Dolz}.
\newblock {Few-Shot Segmentation Without Meta-Learning: A Good Transductive
  Inference Is All You Need?}
\newblock {\em arXiv e-prints}, page arXiv:2012.06166, Dec. 2020.

\bibitem{deeplab}
Liang-Chieh Chen, George Papandreou, Iasonas Kokkinos, Kevin Murphy, and Alan~L
  Yuille.
\newblock Deeplab: Semantic image segmentation with deep convolutional nets,
  atrous convolution, and fully connected crfs.
\newblock {\em IEEE transactions on pattern analysis and machine intelligence},
  40(4):834—848, April 2018.

\bibitem{deeplabv3}
Liang-Chieh Chen, Yukun Zhu, George Papandreou, Florian Schroff, and Hartwig
  Adam.
\newblock Encoder-decoder with atrous separable convolution for semantic image
  segmentation.
\newblock In Vittorio Ferrari, Martial Hebert, Cristian Sminchisescu, and Yair
  Weiss, editors, {\em Computer Vision -- ECCV 2018}, pages 833--851, Cham,
  2018. Springer International Publishing.

\bibitem{selfsupsimclr}
Ting Chen, Simon Kornblith, Mohammad Norouzi, and Geoffrey Hinton.
\newblock A simple framework for contrastive learning of visual
  representations.
\newblock In Hal~Daumé III and Aarti Singh, editors, {\em Proceedings of the
  37th International Conference on Machine Learning}, volume 119 of {\em
  Proceedings of Machine Learning Research}, pages 1597--1607. PMLR, 13--18 Jul
  2020.

\bibitem{selfsupsimclr2}
Ting {Chen}, Simon {Kornblith}, Kevin {Swersky}, Mohammad {Norouzi}, and
  Geoffrey~E. {Hinton}.
\newblock Big self-supervised models are strong semi-supervised learners.
\newblock In {\em Advances in Neural Information Processing Systems},
  volume~33, pages 22243--22255, 2020.

\bibitem{moco}
Xinlei Chen, Haoqi Fan, Ross~B. Girshick, and Kaiming He.
\newblock Improved baselines with momentum contrastive learning.
\newblock {\em CoRR}, abs/2003.04297, 2020.

\bibitem{878f4ef5fd274e93b8b3f9f46b1c3872}
{Ming Ming} Cheng, {Guo Xin} Zhang, {Niloy J.} Mitra, Xiaolei Huang, and {Shi
  Min} Hu.
\newblock Global contrast based salient region detection.
\newblock In {\em 2011 IEEE Conference on Computer Vision and Pattern
  Recognition, CVPR 2011}, Proceedings of the IEEE Computer Society Conference
  on Computer Vision and Pattern Recognition, pages 409--416, United States,
  2011. IEEE Computer Society.

\bibitem{selfsupefros}
Carl Doersch, Abhinav Gupta, and Alexei~A. Efros.
\newblock Unsupervised visual representation learning by context prediction.
\newblock In {\em 2015 IEEE International Conference on Computer Vision
  (ICCV)}, pages 1422--1430, 2015.

\bibitem{dong2018}
Nanqing Dong and E. Xing.
\newblock Few-shot semantic segmentation with prototype learning.
\newblock In {\em BMVC}, 2018.

\bibitem{pascaldataset}
M. Everingham, S. Eslami, L. Gool, Christopher K.~I. Williams, J. Winn, and
  Andrew Zisserman.
\newblock The pascal visual object classes challenge: A retrospective.
\newblock {\em International Journal of Computer Vision}, 111:98--136, 2014.

\bibitem{super-pixel-algo}
Pedro~F. Felzenszwalb and Daniel~P. Huttenlocher.
\newblock Efficient graph-based image segmentation.
\newblock {\em International Journal of Computer Vision}, 59(2):167--181, 2004.

\bibitem{danet}
J. Fu, J. Liu, Haijie Tian, Z. Fang, and H. Lu.
\newblock Dual attention network for scene segmentation.
\newblock {\em 2019 IEEE/CVF Conference on Computer Vision and Pattern
  Recognition (CVPR)}, pages 3141--3149, 2019.

\bibitem{laplacian}
Golnaz Ghiasi and Charless~C. Fowlkes.
\newblock Laplacian pyramid reconstruction and refinement for semantic
  segmentation.
\newblock In Bastian Leibe, Jiri Matas, Nicu Sebe, and Max Welling, editors,
  {\em Computer Vision -- ECCV 2016}, pages 519--534, Cham, 2016. Springer
  International Publishing.

\bibitem{sbd}
Bharath Hariharan, Pablo Arbel{\'a}ez, Ross Girshick, and Jitendra Malik.
\newblock Simultaneous detection and segmentation.
\newblock In David Fleet, Tomas Pajdla, Bernt Schiele, and Tinne Tuytelaars,
  editors, {\em Computer Vision -- ECCV 2014}, pages 297--312, Cham, 2014.
  Springer International Publishing.

\bibitem{resnet101}
Kaiming He, Xiangyu Zhang, Shaoqing Ren, and Jian Sun.
\newblock Deep residual learning for image recognition.
\newblock {\em CoRR}, abs/1512.03385, 2015.

\bibitem{hsu2019unsupervised}
Kyle Hsu, Sergey Levine, and Chelsea Finn.
\newblock Unsupervised learning via meta-learning.
\newblock In {\em International Conference on Learning Representations}, 2019.

\bibitem{ccnet}
Zilong Huang, Xinggang Wang, Lichao Huang, Chang Huang, Yunchao Wei, Humphrey
  Shi, and Wenyu Liu.
\newblock Ccnet: Criss-cross attention for semantic segmentation.
\newblock {\em 2019 IEEE/CVF International Conference on Computer Vision
  (ICCV)}, pages 603--612, 2019.

\bibitem{iic}
Xu Ji, A. Vedaldi, and Jo{\~a}o~F. Henriques.
\newblock Invariant information clustering for unsupervised image
  classification and segmentation.
\newblock {\em 2019 IEEE/CVF International Conference on Computer Vision
  (ICCV)}, pages 9864--9873, 2019.

\bibitem{khodadadeh2019unsupervised}
Siavash Khodadadeh, Ladislau B{\"{o}}l{\"{o}}ni, and Mubarak Shah.
\newblock Unsupervised meta-learning for few-shot image and video
  classification.
\newblock {\em CoRR}, abs/1811.11819, 2018.

\bibitem{selfsuprotation}
Nikos Komodakis and Spyros Gidaris.
\newblock {Unsupervised representation learning by predicting image rotations}.
\newblock In {\em {International Conference on Learning Representations
  (ICLR)}}, Vancouver, Canada, Apr. 2018.

\bibitem{selfsuporder1}
Hsin-Ying Lee, Jia-Bin Huang, Maneesh~Kumar Singh, and Ming-Hsuan Yang.
\newblock Unsupervised representation learning by sorting sequence.
\newblock In {\em IEEE International Conference on Computer Vision}, 2017.

\bibitem{adaptiveprtotypelearning}
Gen Li, Varun Jampani, Laura Sevilla{-}Lara, Deqing Sun, Jonghyun Kim, and
  Joongkyu Kim.
\newblock Adaptive prototype learning and allocation for few-shot segmentation.
\newblock {\em CoRR}, abs/2104.01893, 2021.

\bibitem{refinenet}
Guosheng Lin, A. Milan, Chunhua Shen, and I. Reid.
\newblock Refinenet: Multi-path refinement networks for high-resolution
  semantic segmentation.
\newblock {\em 2017 IEEE Conference on Computer Vision and Pattern Recognition
  (CVPR)}, pages 5168--5177, 2017.

\bibitem{coco}
Tsung-Yi Lin, Michael Maire, Serge Belongie, James Hays, Pietro Perona, Deva
  Ramanan, Piotr Doll{\'a}r, and C.~Lawrence Zitnick.
\newblock Microsoft coco: Common objects in context.
\newblock In David Fleet, Tomas Pajdla, Bernt Schiele, and Tinne Tuytelaars,
  editors, {\em Computer Vision -- ECCV 2014}, pages 740--755, Cham, 2014.
  Springer International Publishing.

\bibitem{parsenet}
W. Liu, Andrew Rabinovich, and A. Berg.
\newblock Parsenet: Looking wider to see better.
\newblock {\em ArXiv}, abs/1506.04579, 2015.

\bibitem{ppnet}
Yongfei Liu, Xiangyi Zhang, Songyang Zhang, and Xuming He.
\newblock Part-aware prototype network for few-shot semantic segmentation.
\newblock {\em CoRR}, abs/2007.06309, 2020.

\bibitem{hsnet}
Juhong Min, Dahyun Kang, and Minsu Cho.
\newblock Hypercorrelation squeeze for few-shot segmentation.
\newblock {\em CoRR}, abs/2104.01538, 2021.

\bibitem{selfsuporder2}
Ishan Misra, C.~Lawrence Zitnick, and Martial Hebert.
\newblock {Shuffle and Learn: Unsupervised Learning using Temporal Order
  Verification}.
\newblock In {\em ECCV}, 2016.

\bibitem{DeepUSPS}
Duc~Tam Nguyen, Maximilian Dax, Chaithanya~Kumar Mummadi, Thi{-}Phuong{-}Nhung
  Ngo, Thi Hoai~Phuong Nguyen, Zhongyu Lou, and Thomas Brox.
\newblock Deepusps: Deep robust unsupervised saliency prediction with
  self-supervision.
\newblock {\em CoRR}, abs/1909.13055, 2019.

\bibitem{fwb}
Khoi Nguyen and Sinisa Todorovic.
\newblock Feature weighting and boosting for few-shot segmentation.
\newblock In {\em 2019 IEEE/CVF International Conference on Computer Vision
  (ICCV)}, pages 622--631, 2019.

\bibitem{selfsupjigsaw}
Mehdi Noroozi and Paolo Favaro.
\newblock Unsupervised learning of visual representations by solving jigsaw
  puzzles.
\newblock In {\em ECCV}, 2016.

\bibitem{autounsupseg}
Yassine Ouali, Celine Hudelot, and Myriam Tami.
\newblock Autoregressive unsupervised image segmentation.
\newblock In {\em Proceedings of the European Conference on Computer Vision
  (ECCV)}, August 2020.

\bibitem{self-supervised_fss}
Cheng Ouyang, Carlo Biffi, Chen Chen, Turkay Kart, Huaqi Qiu, and Daniel
  Rueckert.
\newblock Self-supervision with superpixels: Training few-shot medical image
  segmentation without annotation.
\newblock {\em CoRR}, abs/2007.09886, 2020.

\bibitem{selfsuppainting}
Deepak Pathak, Philipp Kr\"ahenb\"uhl, Jeff Donahue, Trevor Darrell, and Alexei
  Efros.
\newblock Context encoders: Feature learning by inpainting.
\newblock In {\em Computer Vision and Pattern Recognition ({CVPR})}, 2016.

\bibitem{BasNet}
Xuebin Qin, Zichen Zhang, Chenyang Huang, Chao Gao, Masood Dehghan, and Martin
  Jagersand.
\newblock Basnet: Boundary-aware salient object detection.
\newblock In {\em Proceedings of the IEEE/CVF Conference on Computer Vision and
  Pattern Recognition (CVPR)}, June 2019.

\bibitem{rakelly2018}
Kate Rakelly, Evan Shelhamer, Trevor Darrell, Alexei~A. Efros, and Sergey
  Levine.
\newblock Conditional networks for few-shot semantic segmentation.
\newblock In {\em ICLR}, 2018.

\bibitem{unet}
Olaf Ronneberger, Philipp Fischer, and Thomas Brox.
\newblock U-net: Convolutional networks for biomedical image segmentation.
\newblock In Nassir Navab, Joachim Hornegger, William~M. Wells, and
  Alejandro~F. Frangi, editors, {\em Medical Image Computing and
  Computer-Assisted Intervention -- MICCAI 2015}, pages 234--241, Cham, 2015.
  Springer International Publishing.

\bibitem{imagenet}
Olga Russakovsky, Jia Deng, Hao Su, Jonathan Krause, Sanjeev Satheesh, Sean Ma,
  Zhiheng Huang, Andrej Karpathy, Aditya Khosla, Michael~S. Bernstein,
  Alexander~C. Berg, and Fei{-}Fei Li.
\newblock Imagenet large scale visual recognition challenge.
\newblock {\em CoRR}, abs/1409.0575, 2014.

\bibitem{oslsm}
Amirreza Shaban, Shray Bansal, Zhen Liu, Irfan Essa, and Byron Boots.
\newblock One-shot learning for semantic segmentation.
\newblock 2017.

\bibitem{fcn}
Evan Shelhamer, J. Long, and Trevor Darrell.
\newblock Fully convolutional networks for semantic segmentation.
\newblock {\em IEEE Transactions on Pattern Analysis and Machine Intelligence},
  39:640--651, 2017.

\bibitem{amp}
Mennatullah Siam and Boris~N. Oreshkin.
\newblock Adaptive masked weight imprinting for few-shot segmentation.
\newblock {\em CoRR}, abs/1902.11123, 2019.

\bibitem{pfenet}
Zhuotao Tian, Hengshuang Zhao, Michelle Shu, Zhicheng Yang, Ruiyu Li, and Jiaya
  Jia.
\newblock Prior guided feature enrichment network for few-shot segmentation.
\newblock {\em CoRR}, abs/2008.01449, 2020.

\bibitem{maskcontrast}
Wouter Van~Gansbeke, Simon Vandenhende, Stamatios Georgoulis, and Luc Van~Gool.
\newblock Unsupervised semantic segmentation by contrasting object mask
  proposals.
\newblock In {\em International Conference on Computer Vision}, 2021.

\bibitem{democratic}
Haochen Wang, Xudong Zhang, Yutao Hu, Yandan Yang, Xianbin Cao, and Xiantong
  Zhen.
\newblock Few-shot semantic segmentation with democratic attention networks.
\newblock In {\em ECCV}, 2020.

\bibitem{panet}
K. Wang, J. Liew, Yingtian Zou, Daquan Zhou, and Jiashi Feng.
\newblock Panet: Few-shot image semantic segmentation with prototype alignment.
\newblock {\em 2019 IEEE/CVF International Conference on Computer Vision
  (ICCV)}, pages 9196--9205, 2019.

\bibitem{scaleawaregnn}
Guo-Sen Xie, Jie Liu, Huan Xiong, and Ling Shao.
\newblock Scale-aware graph neural network for few-shot semantic segmentation.
\newblock In {\em Proceedings of the IEEE/CVF Conference on Computer Vision and
  Pattern Recognition (CVPR)}, pages 5475--5484, June 2021.

\bibitem{cyclicmemory}
Guo-Sen Xie, Huan Xiong, Jie Liu, Yazhou Yao, and Ling Shao.
\newblock Few-shot semantic segmentation with cyclic memory network.
\newblock In {\em Proceedings of the IEEE/CVF International Conference on
  Computer Vision (ICCV)}, pages 7293--7302, October 2021.

\bibitem{ppm}
Boyu Yang, Chang Liu, Bohao Li, Jianbin Jiao, and Qixiang Ye.
\newblock Prototype mixture models for few-shot semantic segmentation.
\newblock In {\em ECCV}, 2020.

\bibitem{mininglatentclasses}
Lihe Yang, Wei Zhuo, Lei Qi, Yinghuan Shi, and Yang Gao.
\newblock Mining latent classes for few-shot segmentation.
\newblock {\em CoRR}, abs/2103.15402, 2021.

\bibitem{dilated}
Fisher Yu and Vladlen Koltun.
\newblock Multi-scale context aggregation by dilated convolutions.
\newblock In Yoshua Bengio and Yann LeCun, editors, {\em 4th International
  Conference on Learning Representations, {ICLR} 2016, San Juan, Puerto Rico,
  May 2-4, 2016, Conference Track Proceedings}, 2016.

\bibitem{convnet-visu}
Matthew~D. Zeiler and Rob Fergus.
\newblock Visualizing and understanding convolutional networks.
\newblock {\em CoRR}, abs/1311.2901, 2013.

\bibitem{SCL}
Bingfeng Zhang, Jimin Xiao, and Terry Qin.
\newblock Self-guided and cross-guided learning for few-shot segmentation.
\newblock {\em CoRR}, abs/2103.16129, 2021.

\bibitem{pgnet}
Chi Zhang, Guosheng Lin, Fayao Liu, Jiushuang Guo, Qingyao Wu, and Rui Yao.
\newblock Pyramid graph networks with connection attentions for region-based
  one-shot semantic segmentation.
\newblock In {\em Proceedings of the IEEE/CVF International Conference on
  Computer Vision (ICCV)}, October 2019.

\bibitem{canet}
Chi Zhang, Guosheng Lin, Fayao Liu, Rui Yao, and Chunhua Shen.
\newblock Canet: Class-agnostic segmentation networks with iterative refinement
  and attentive few-shot learning.
\newblock {\em 2019 IEEE/CVF Conference on Computer Vision and Pattern
  Recognition (CVPR)}, pages 5212--5221, 2019.

\bibitem{contextencoding}
H. Zhang, K. Dana, J. Shi, Zhongyue Zhang, Xiaogang Wang, A. Tyagi, and Amit
  Agrawal.
\newblock Context encoding for semantic segmentation.
\newblock {\em 2018 IEEE/CVF Conference on Computer Vision and Pattern
  Recognition}, pages 7151--7160, 2018.

\bibitem{cooccur}
H. Zhang, Chenguang Wang, and Junyuan Xie.
\newblock Co-occurrent features in semantic segmentation.
\newblock {\em 2019 IEEE/CVF Conference on Computer Vision and Pattern
  Recognition (CVPR)}, pages 548--557, 2019.

\bibitem{pspnet}
Hengshuang Zhao, J. Shi, Xiaojuan Qi, Xiaogang Wang, and J. Jia.
\newblock Pyramid scene parsing network.
\newblock {\em 2017 IEEE Conference on Computer Vision and Pattern Recognition
  (CVPR)}, pages 6230--6239, 2017.

\bibitem{psanet}
Hengshuang Zhao, Yi Zhang, Shu Liu, Jianping Shi, Chen~Change Loy, Dahua Lin,
  and Jiaya Jia.
\newblock Psanet: Point-wise spatial attention network for scene parsing.
\newblock In {\em Proceedings of the European Conference on Computer Vision
  (ECCV)}, September 2018.

\bibitem{selfsuptuning}
Kai {Zhu}, Wei {Zhai}, Zheng-Jun {Zha}, and Yang {Cao}.
\newblock {Self-Supervised Tuning for Few-Shot Segmentation}.
\newblock {\em arXiv e-prints}, page arXiv:2004.05538, Apr. 2020.

\end{thebibliography}
}

\end{document}